\documentclass[10pt,twocolumn,letterpaper]{article}

\usepackage{iccv}
\usepackage{times}
\usepackage{epsfig}
\usepackage{graphicx}
\usepackage{amsmath}
\usepackage{amssymb}

\usepackage{booktabs}
\usepackage{multirow}
\usepackage{multicol}
\usepackage{subcaption}

\usepackage[pagebackref=true,breaklinks=true,letterpaper=true,colorlinks,bookmarks=false]{hyperref}

\iccvfinalcopy 

\def\iccvPaperID{17} 

\def\x{\times}
\ificcvfinal\fi

\begin{document}
	
	\title{LUAI Challenge 2021 on Learning to Understand Aerial Images}
	
	
	\author{
		Gui-Song Xia$^1$, Jian Ding$^1$, Ming Qian$^1$, Nan Xue$^1$, Jiaming Han$^1$, Xiang Bai$^2$, Michael Ying Yang$^3$, \\ Shengyang Li$^4$, Serge Belongie$^5$, Jiebo Luo$^6$, Mihai Datcu$^{7,8}$, Marcello Pelillo$^9$, Liangpei Zhang$^1$, \\ Qiang Zhou$^{10}$, Chao-hui Yu$^{10}$, Kaixuan Hu$^{11}$, Yingjia Bu$^{11}$, Wenming Tan$^{11}$, Zhe  Yang$^{10}$, \\Wei Li$^{10}$, Shang Liu$^{10}$, Jiaxuan Zhao$^{12}$, Tianzhi Ma$^{12}$, Zi-han Gao$^{12}$, Lingqi Wang$^{12}$, \\ Yi Zuo$^{12}$, Licheng Jiao$^{12}$, Chang  Meng$^{12}$, Hao  Wang$^{12}$, Jiahao Wang$^{12}$, Yiming Hui$^{12}$, \\ Zhuojun Dong$^{12}$, Jie Zhang$^{12}$, Qianyue Bao$^{12}$, Zixiao Zhang$^{12}$, Fang Liu$^{12}$ \\
		\\
		$^1${\em Wuhan University, China.} \\
		$^2${\em Huazhong University of Science and Technology, China.}\\
		$^3${\em University of Twente, Netherlands.}\\
		$^4${\em Chinese Academy of Sciences, Beijing, China}. \\
		$^5${\em Cornell Tech and Cornell University, United States.} \\
		$^6${\em University of Rochester, United States.}\\
		$^7${\em German Aerospace Center (DLR), Germany.} \\ 
		$^8${\em University POLITEHNICA of Bucharest, Romania. } \\ 
		$^9${\em Computer Science, Ca’ Foscari University of Venice, Italy.} \\
		$^{10}${\em Alibaba Group, China.} \\
		$^{11}${\em Hikvision Research Institute} \\
		$^{12}${\em Xidian University, China.}
	}
	
	\maketitle
	\ificcvfinal\thispagestyle{empty}\fi
	
	\begin{abstract}
		This report summarizes the results of Learning to Understand Aerial Images (LUAI) 2021 challenge held on ICCV’2021, which focuses on object detection and semantic segmentation in aerial images. Using DOTA-v2.0~\cite{ding2021object} and GID-15~\cite{GID2020} datasets, this challenge proposes three tasks for oriented object detection, horizontal object detection, and semantic segmentation of common categories in aerial images. This challenge received a total of 146 registrations on the three tasks. 
		Through the challenge, we hope to draw attention from a wide range of communities and call for more efforts on the problems of learning to understand aerial images.
	\end{abstract}
	
	\section{Introduction}
	
	Earth vision, also known as {\em Earth Observation and Remote Sensing}, targets to understand large-scale scenes on the Earth's surface with aerial images taken from overhead view, which provides a new way to understand our physical world and benefits many applications, {\em e.g.}, urban management and planning, precise agriculture, emergency rescue and disaster relief.

	Recently, the problems in Earth vision span from plane detection~\cite{rareplane}, ship detection~\cite{HRSC2016}, vehicle detection~\cite{DLR3KMunichVehicle}, building extraction~\cite{SpaceNet_MVOI}, road extraction~\cite{DeepGlobe}, and changing detection~\cite{ABCD}, etc. Most of these applications can be considered as special cases of object detection~\cite{DOTA,ICPRODAI,s2anet,maskobb,RRD,GlidVertex}, semantic segmentation~\cite{GID2020}, and instance segmentation~\cite{ISOP,iSAID} of aerial images. 
	
	To advance methods for learning to understand aerial images, we propose a competition that focuses on object detection and semantic segmentation for common categories on an ICCV workshop\footnote{\url{https://captain-whu.github.io/LUAI2021/challenge.html}}. As known to all, large-scale, well-annotated, and diverse datasets are crucial for learning-based algorithms. Therefore, for object detection in aerial images, this competition uses a new large-scale benchmark database, \ie, DOTA-v2.0~\cite{ding2021object}\footnote{\url{https://captain-whu.github.io/DOTA}.}. The dataset contains 11,268 large-scale images (the maximum size is 20,000), 18 categories, and $1,793,658$ instances, each of which is labeled by an arbitrary (8 d.o.f.) oriented bounding box. 
	It is worth noticing that the competition tasks on object detection follows two previous {\em object detection in aerial images} (ODAI) challenges held on ICPR'2018~\cite{ICPRODAI} and CVPR'2019~\cite{doai2019}, using DOTA-v1.0~\cite{DOTA} and DOTA-v1.5~\cite{doai2019}, respectively.
	For semantic segmentation, we propose a new dataset, named as Gaofen Image Dataset with 15 categories (GID-15~\cite{GID2020}\footnote{\url{https://captain-whu.github.io/GID15/}}). GID-15 contains 150 pixel-level annotated GF-2 images in 15 categories. Based on the two datasets, we propose three competition tasks, namely oriented object detection, horizontal object detection, and semantic segmentation.
	
	
	Through the dataset and competition tasks, we aim to draw attention from a wide range of communities and call for more future research and efforts on the problems of object detection and semantic segmentation in aerial images. 
	We believe the competition and workshop will not only promote the development of algorithms in Earth vision, but also pose interesting algorithmic questions to the computer vision community.

	\section{Datasets}
	In this section, we present the statics and properties of the two datasets for object detection and semantic segmentation respectively.
	
	\subsection{Object Detection}
\begin{figure*}[t!]
		\begin{center}
			\includegraphics[width=0.8\linewidth]{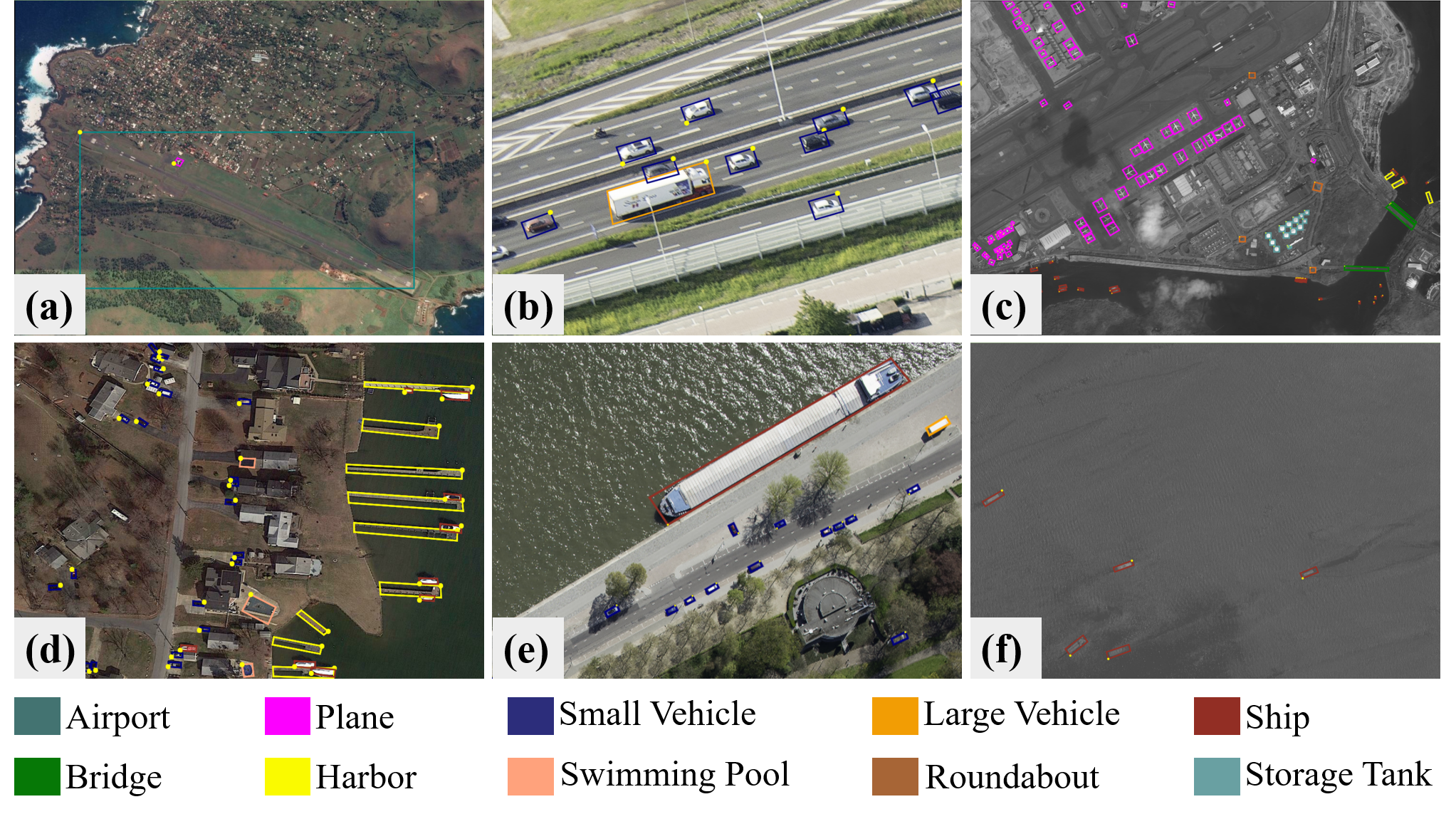}
		\end{center}
		\vspace{-4mm}
		\caption{Examples of images and their corresponding annotations in DOTA-v2.0. (a) and (d) are Google Earth images. (b) and (e) are Airborne images taken by CycloMedia. (c) and (f) are panchromatic band of GF-2 satellite images. Note that we only show part of the categories of DOTA-v2.0. To view the details of the full DOTA-v2.0, you can refer to ~\cite{ding2021object}.}
		\label{fig:dota2}
		\vspace{-2mm}
	\end{figure*}

	The object detection track is based on DOTA-v2.0~\cite{ding2021object}, which collects images from Google Earth, GF- 2 Satellite, and aerial images. The \textit{domain shifts} among these different image sources make it possible to develop \textit{domain-robust} object detectors, which also have practical value. There are 18 common categories, 11,268 images and 1,793,658 instances in DOTA-v2.0. All the objects in DOTA-v2.0 are annotated with oriented bounding boxes. These images and oriented annotations provide a rich resource for researches on rotation-invariant object detection. DOTA-v2.0 are split into training (1,830 images), validation (593 images), test-dev (2792 images), and test-challenge (6053 images). The evaluation in this challenge is on the \textbf{test-challenge}. We do not release the ground truths. The scores can be obtained by submitting results to the evaluation server. The annotated examples for can be found in Fig.~\ref{fig:dota2}.   
	
	
	\subsection{Semantic Segmentation}
\begin{figure*}[t!]
		\begin{center}
			\includegraphics[width=0.97\linewidth]{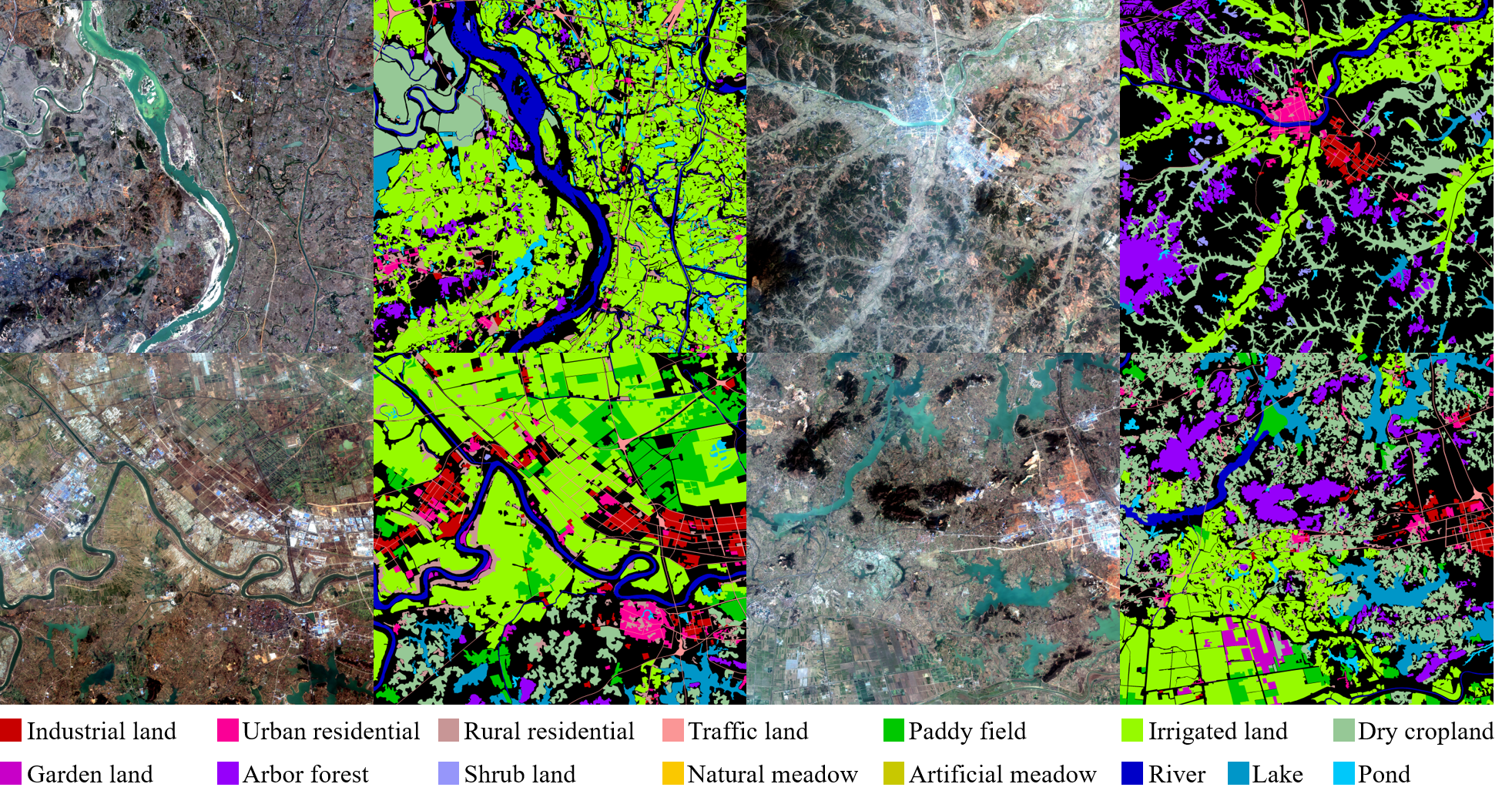}
		\end{center}
		\vspace{-4mm}
		\caption{Examples of images and their corresponding annotations in GID-15.}
		\label{fig:gid}
		\vspace{-2mm}
	\end{figure*}

	The semantic segmentation track is based on Gaofen Image Dataset with 15 categories (GID-15)~\cite{GID2020}, which is a new large-scale land-cover dataset. GID-15 has many superiorities over the exsiting land-cover dataset owing to its large coverage, wide distribution, and high spatial resolution. The large-scale remote sensing semantic segmentation dataset contains 150 pixel-level annotated GF-2 images in 15 categories. The image sizes of the GF-2 images are $7,200\times6,800$. The examples of annotated images in GID-15 can be found in Fig.~\ref{fig:gid}.
	
	\section{Challenge Tasks}
	Using the above mentioned DOTA-v2.0~\cite{ding2021object} and GID-15, we propose three tasks, namely oriented object detection,  horizontal object detection, and semantic segmentation. In what follows, we provide the details of outputs and evaluation metrics of each task.
	\subsection{Task1 - Oriented Object Detection}
	The target of this task is to locate objects with oriented bounding boxes (OBBs) and give their semantic labels. The OBBs here are quadrilaterals $\{(x_i, y_i)|i=1,2,3,4\}$. We adopt the mean average precision (mAP) as the evaluation metric, which follows the PASCAL VOC~\cite{PASCALVOC} except that the IoU calculation is performed between OBBs.
	
	\subsection{Task2 - Horizontal Object Detection}
	The target of this task is to locate objects with horizontal bounding boxes and give their semantic labels. The HBBs here are rectangles $(x_{min}, y_{min}, x_{max}, y_{max})$, which are generated by calculating axis-aliened bounding boxes of OBBs. We adopt the mean average precision (mAP) as the evaluation metric, which is exactly the same as PASCAL VOC~\cite{PASCALVOC}.
	
	\subsection{Task3 - Semantic Segmentation}
	The target of this task is to predict the semantic labels for all the pixels in aerial images. We adopt the mean intersection over union (mIoU)~\cite{PASCALVOC} as the evaluation metric.
	\section{Organization}
	
	The registration of this competition starts on July 5, 2021. After registration, the participants can download the images of the train/val/test set and ground truths of the train/val set. Since we do not release the ground truths of these tasks, the participants need to submit their results to our evaluation server for automatic evaluation. If the evaluation process succeeds, the results will be sent to the registered emails. Each team is allowed to submit only once a day to the server during the challenge to avoid overfitting. The deadline for challenge submission is Aug 15, 2021.
	
	There are 84 and 62 teams registered at the DOTA-v2.0 evaluation server and the GID-15 evaluation server, respectively. 
	The teams come from universities, research institutes, and tech companies, such as Alibaba, Tsinghua University, Hikvision Research Institute, Xidian University, Northwestern Polytechnical University (NWPU), and Satrec Initiative.
	
	There are 459, 219, and 374 submissions for task1, task2, and task3, respectively. If one team has multiple submissions, only the submission with the highest score is kept. Each team is asked to submit a description of their method. Otherwise, the submissions are not valid. 
	
	
	\section{Submissions and Results}
	The mAP of two object detection tasks and mIoU of semantic segmentation task are calculated by our evaluation server automatically. The Tab.~\ref{tab:task1}, Tab.~\ref{tab:task2}, and Tab.~\ref{tab:task3} summarize the valid results of each team on three tasks, respectively. 
	
	\subsection{Baseline Results}
	For reference, we implement and evaluate baseline algorithms for each task. For Task1 and Task2, we adopt RoI Transformer, R50-\textbf{FPN}, \textbf{1$\x$} schedule~\cite{RoITransformer} and ReDet, ReR50-\textbf{ReFPN}, \textbf{1$\x$} schedule~\cite{ReDet} as our baselines. We follow the single scale default settings in AerialDetection Benchmark~\cite{ding2021object}\footnote{\url{https://github.com/dingjiansw101/AerialDetection}}. For Task3, we adopt UperNet~\cite{upernet} with Swin-S~\cite{swin} as our baseline. When training, we only use training data(100 images) and no validation. We change the input size to 1024$\x$1024. Then we ignore the background area when calculating loss. When testing an image, we adopt a slide inference strategy. The patch size is 1024$\x$1024, and stride is 700$\x$700 in this process. Besides, we follow the default setting in Swin Transformer when training on Cityscapes\cite{Cordts2015Cvprw} dataset.
	
	\begin{table*}[t!]
		\caption{Results of valid submissions in Task1. The abbreviations for categories are defined as: {\em BD--Baseball diamond, GTF--Ground field track, SV--Small vehicle,  LV--Large vehicle, TC--Tennis court, BC--Basketball court, SC--Storage tank, SBF--Soccer-ball field, RA--Roundabout, SP--Swimming pool, HC--Helicopter, CC--Container Crane, Air--Airport, HP--Helipad}.
		}
		\centering
		\vspace{-3mm}
		\resizebox{1.0\linewidth}{!}{
			\begin{tabular}{cccccccccccccccccccc}
				\hline
				Team               & mAP   & PL    & BD    & Bridge & GTF   & SV    & LV    & Ship  & TC    & BC    & ST    & SBF   & RA    & Harbor & SP    & HC    & CC    & Airport & HP    \\ \hline
				Alibaba\_AI\_Earth & 58.97 & 87.34 & 62.82 & 46.26  & 76.89 & 68.9  & 42.64 & 83.98 & 63.38 & 50.01 & 78.98 & 51.75 & 58.81 & 47.57  & 62.98 & 14.02 & 53.88 & 88.41   & 22.79 \\
				Tsinghua\_cc       & 56.03 & 86.9  & 61.71 & 45.42  & 74.96 & 62.5  & 39.48 & 81.41 & 60.88 & 45.27 & 77.61 & 47.56 & 57.88 & 44.8   & 61.87 & 11.62 & 49.27 & 79.88   & 19.5  \\
				HIK\_HOW           & 55.93 & 86    & 61.88 & 46.78  & 76.03 & 55.19 & 38.01 & 77.57 & 59.42 & 45.57 & 74.17 & 50.4  & 55.8  & 46.5   & 58.06 & 10.6  & 53.07 & 85.26   & 26.37 \\
				Beaters            & 54.59 & 85.18 & 58.73 & 44.44  & 75.03 & 56.64 & 34.75 & 76.79 & 55.51 & 43.96 & 73.71 & 49.27 & 55.36 & 43.48  & 58.05 & 18.29 & 47.81 & 82.1    & 23.52 \\
				LMY-XiaoXiao       & 50.08 & 85.91 & 57.39 & 43.55  & 75.06 & 46.2  & 32.87 & 76.23 & 57.39 & 46.21 & 50.85 & 49.91 & 46.15 & 43.03  & 53.58 & 9.75  & 47.34 & 70.22   & 9.74  \\
				SJTU\_Pingan\_KYZ  & 47.98 & 83.91 & 58.45 & 38.93  & 73    & 52.7  & 27.52 & 66.21 & 54.97 & 42.07 & 73.13 & 43.97 & 52.56 & 31.66  & 53.64 & 13.97 & 28.63 & 46.06   & 22.28 \\
				\_pure\_white\_    & 45.23 & 79.15 & 56.89 & 41.26  & 73.64 & 26.73 & 24.66 & 59.45 & 56.82 & 42.81 & 59    & 47.53 & 42.82 & 40.23  & 46.78 & 5.5   & 31.55 & 70.22   & 9.01  \\
				Im2hard            & 44.37 & 77.29 & 52.17 & 38.16  & 70.99 & 26.52 & 21.65 & 59.11 & 52.22 & 45.82 & 63.4  & 37.81 & 45.47 & 34.81  & 45.15 & 1.49  & 45.38 & 64.3    & 16.98 \\
				hakjinlee          & 42.39 & 77.27 & 52.29 & 39.18  & 70.86 & 26.25 & 20.26 & 58.63 & 46.55 & 36.99 & 58.86 & 41.22 & 45.31 & 34.93  & 48.27 & 2.71  & 21.02 & 71.21   & 11.28 \\
				xyh                & 31.24 & 61.29 & 40.99 & 19.58  & 61.19 & 25.95 & 14.5  & 48.01 & 25.6  & 4.71  & 52.02 & 29.55 & 36    & 9.96   & 38.05 & 0     & 10.27 & 68.66   & 15.95 \\
				phale              & 18.55 & 51.71 & 29.2  & 12.64  & 45.57 & 16.75 & 9.09  & 22.39 & 14.58 & 9.09  & 34.68 & 15.92 & 35.12 & 11     & 0.64  & 0     & 0     & 23.11   & 2.46  \\ \hline
				RoI Trans.~\cite{RoITransformer} (baseline)  & 38.34 & 70.38 & 45.46 & 32.72  & 67.71 & 26.60 & 21.06 & 51.22 & 43.15 & 30.42 & 59.46 & 34.57 & 43.22 & 28.41  & 43.42 & 1.82  & 18.28 & 62.30   & 9.86  \\
				ReDet~\cite{ReDet} (baseline)            & 40.05 & 73.93 & 48.85 & 35.58  & 68.62 & 26.51 & 19.24 & 56.89 & 48.73 & 32.58 & 59.89 & 35.82 & 44.36 & 32.21  & 37.41 & 3.20  & 18.97 & 69.09   & 9.03 \\ \hline
			\end{tabular}\label{tab:task1}						
		}
		\vspace{-1mm}
	\end{table*}

	\begin{table*}[t!]
		\caption{Results of valid submissions in Task2. The abbreviations are the same as in Tab.~\ref{tab:task1}.
		}
		\centering
		\vspace{-3mm}
		\resizebox{1.0\linewidth}{!}{
			\begin{tabular}{cccccccccccccccccccc}
				\hline
				Team               & mAP   & PL    & BD    & Bridge & GTF   & SV    & LV    & Ship  & TC    & BC    & ST    & SBF   & RA    & Harbor & SP    & HC    & CC    & Airport & HP    \\ \hline
				Alibaba\_AI\_Earth & 60.31 & 87.47 & 63.6  & 49.85  & 76.86 & 75.25 & 43.5  & 85.35 & 63.78 & 50.12 & 80.38 & 51.76 & 59.24 & 52.95  & 64.57 & 14.02 & 55.65 & 88.3    & 22.85 \\
				HIK\_HOW           & 57.6  & 86.28 & 62.29 & 49.35  & 75.96 & 65.33 & 40.84 & 78.38 & 60.33 & 45.71 & 74.53 & 50.37 & 54.9  & 53.01  & 60.24 & 10.6  & 56.15 & 85.97   & 26.48 \\
				Tsinghua\_cc       & 57.57 & 87.03 & 62.91 & 47.28  & 75.02 & 70.61 & 40.24 & 84.44 & 62.03 & 45.27 & 79.19 & 47.43 & 58.1  & 49.15  & 63.62 & 13.19 & 51.26 & 79.93   & 19.5  \\
				Beaters            & 55.86 & 85.5  & 58.89 & 46.72  & 75.05 & 65.71 & 36.05 & 77.87 & 55.58 & 44.01 & 74.37 & 49.33 & 55.7  & 47.68  & 59.36 & 18.29 & 49.63 & 82.14   & 23.67 \\
				xyh                & 32.98 & 61.19 & 40.91 & 23.88  & 60.64 & 26.07 & 15.95 & 50.45 & 29.02 & 5.05  & 52.07 & 29.39 & 36.14 & 18.42  & 38.68 & 0     & 21.51 & 68.69   & 15.65 \\ \hline
				RoI Trans. (baseline)~\cite{RoITransformer}  & 39.34 & 70.47 & 45.84 & 34.70  & 67.68 & 26.68 & 22.38 & 57.74 & 43.17 & 30.56 & 59.73 & 34.67 & 43.36 & 33.68  & 46.24 & 1.82  & 19.71 & 62.50   & 7.13  \\
				ReDet (baseline)~\cite{ReDet}           & 41.06 & 74.37 & 48.86 & 38.01  & 68.61 & 26.60 & 21.29 & 58.80 & 49.07 & 32.79 & 60.10 & 36.08 & 44.59 & 35.91  & 41.03 & 3.20  & 22.95 & 69.15   & 7.71 \\ \hline
			\end{tabular}	\label{tab:task2}
		}
		\vspace{-1mm}
	\end{table*}
	
	\begin{table*}[t!]
		\caption{Results of valid submissions in Task3. The abbreviations for categories are defined as: {\em IDL--industrial land, UR--urban residential, RR--rural residential, TL--traffic land, PF--paddy field, IGL--irrigated land, DC--dry cropland, GP--garden plot, AW--arbor woodland, SL--shrub land, NG--natural grassland, AG--artificial grassland}.
		}
		\centering
		\vspace{-3mm}
		\resizebox{1.0\linewidth}{!}{
			\begin{tabular}{ccccccccccccccccc}
				\hline
				Team              & mIoU  & IDL   & UR    & RR    & TL    & PF    & IGL   & DC    & GP    & AW    & SL    & NG    & AG    & River & Lake  & Pond  \\ \hline
				Alibaba\_AI\_Earth    & 64.54 & 75.63 & 81.11 & 75.51 & 76.58 & 58.59 & 86.66 & 68.55 & 45.19 & 68.51 & 34.03 & 84.66 & 25.3  & 74.09 & 85.16 & 28.48 \\
				lingling              & 58.99 & 68.82 & 77.27 & 70.75 & 60.88 & 64.64 & 82.96 & 66.03 & 18.1  & 71.32 & 35.27 & 64.28 & 17.69 & 71.65 & 85.39 & 29.76 \\
				Go for it             & 58.24 & 71.96 & 77.86 & 71.21 & 65.79 & 57.31 & 82.98 & 61.43 & 29.69 & 70.14 & 27.81 & 70.63 & 21.89 & 62.22 & 78.5  & 24.22 \\
				dong                  & 57.86 & 70.61 & 77.54 & 70.26 & 68.8  & 59.67 & 80.63 & 55.69 & 32.28 & 72.64 & 34.79 & 65.44 & 12.65 & 62.43 & 78.64 & 25.81 \\
				pku\_\_lizhou         & 56.89 & 68.81 & 77.26 & 70.75 & 60.84 & 51.91 & 78.93 & 56.37 & 25.55 & 71.63 & 35.27 & 63.28 & 14.27 & 70.87 & 85.16 & 22.41 \\
				deepblue\_baijieying  & 56.87 & 68.81 & 77.26 & 70.75 & 60.84 & 51.55 & 78.93 & 56.37 & 25.55 & 71.63 & 35.27 & 63.28 & 14.27 & 70.96 & 85.17 & 22.39 \\
				zds                   & 55.83 & 69.38 & 77.34 & 73.6  & 72.83 & 53.68 & 82.2  & 59.8  & 26.46 & 68.12 & 9.68  & 75.35 & 10.2  & 58.87 & 78.68 & 21.33 \\
				SKKU - Automation Lab & 51.04 & 68.93 & 76.57 & 59.83 & 60.67 & 51.01 & 76.88 & 46.47 & 25.15 & 60.47 & 3.24  & 60.67 & 13.87 & 62.11 & 79.63 & 20.03 \\
				yixinzhishui          & 50.14 & 64.77 & 73.86 & 63.03 & 54.71 & 42.75 & 80.22 & 59.46 & 17.88 & 62.6  & 6.8   & 60.85 & 2.86  & 61.92 & 78.1  & 22.32 \\
				zhaosijie             & 48.37 & 65.61 & 74.66 & 56.32 & 52.51 & 52.31 & 76.4  & 49.26 & 17.12 & 59.85 & 5.01  & 57.48 & 3.18  & 60.03 & 80.4  & 15.34 \\
				Amadeus               & 47.46 & 67.08 & 75.1  & 68.77 & 57.32 & 51.1  & 75.22 & 40.64 & 13.05 & 65.39 & 1.71  & 37.93 & 4.21  & 53.74 & 75.09 & 25.52 \\
				DeepBlueAI            & 44.05 & 64.87 & 75.28 & 66.98 & 63.46 & 40.99 & 71.65 & 13.86 & 7.5   & 66.12 & 1.09  & 59.58 & 4.99  & 45.64 & 65.83 & 12.87 \\ \hline
				UperNet + Swin-S~\cite{swin} (baseline)    & 58.35 & 73.03 & 78.68 & 73    & 74.17 & 53.51 & 83.51 & 59.18 & 35.84 & 66.16 & 4.73  & 80.35 & 20.7  & 64.11 & 80.13 & 28.1 \\ \hline 
			\end{tabular}	\label{tab:task3}
		}
		\vspace{-3mm}
	\end{table*}

	\begin{table*}[t!]
		\caption{Ablations of Alibaba\_AI\_Earth in Task1.}
		\vspace{-3mm}
		\centering 
		\resizebox{1\linewidth}{!}{
			\begin{tabular}{lccccccccccccccccccc}
				\hline
				Model                      & mAP   & PL    & BD    & Bridge & GTF   & SV    & LV    & Ship  & TC    & BC    & ST    & SBF   & RA    & Harbor & SP    & HC    & CC    & Airport & HP    \\ \hline
				baseline                   & 46.40 & 77.26 & 50.25 & 42.77  & 71.90 & 28.52 & 27.73 & 66.24 & 57.80 & 40.37 & 67.18 & 41.89 & 46.11 & 38.81  & 44.09 & 2.76  & 43.77 & 69.07   & 14.94 \\
				+single model optimization & 56.86 & 86.20 & 61.92 & 46.16  & 74.94 & 61.94 & 41.57 & 82.93 & 63.28 & 48.11 & 77.63 & 49.42 & 58.40 & 46.19  & 61.90 & 13.89 & 51.72 & 80.00   & 17.22 \\
				+TTA                       & 58.97 & 87.34 & 62.82 & 46.26  & 76.89 & 68.90 & 42.64 & 83.98 & 63.38 & 50.01 & 78.98 & 51.75 & 58.81 & 47.57  & 62.98 & 14.02 & 53.88 & 88.41   & 22.79 \\ \hline
			\end{tabular}
		}
		\label{tbl:results_rbb}
	\end{table*}
	
	\begin{table*}[t!]
		\caption{Ablations of Alibaba\_AI\_Earth in Task2.}
		\vspace{-3mm}
		\centering 
		\resizebox{1\linewidth}{!}{
			\begin{tabular}{lccccccccccccccccccc}
				\hline
				Model                      & mAP   & PL    & BD    & Bridge & GTF   & SV    & LV    & Ship  & TC    & BC    & ST    & SBF   & RA    & Harbor & SP    & HC    & CC    & Airport & HP    \\ \hline
				baseline                   & 47.63 & 79.38 & 54.37 & 43.55  & 73.41 & 32.15 & 29.95 & 67.63 & 59.87 & 43.90 & 50.16 & 49.26 & 35.06 & 47.62  & 53.48 & 2.59  & 51.08 & 73.88   & 9.94  \\
				+single model optimization & 58.25 & 86.38 & 62.61 & 47.87  & 74.96 & 70.40 & 42.42 & 84.95 & 63.48 & 48.05 & 79.37 & 49.26 & 58.83 & 50.61  & 63.73 & 13.89 & 53.61 & 79.99   & 18.02 \\
				+TTA                       & 60.30 & 87.47 & 63.60 & 49.85  & 76.86 & 75.25 & 43.50 & 85.35 & 63.78 & 50.12 & 80.32 & 51.76 & 59.24 & 52.95  & 64.57 & 14.02 & 55.65 & 88.30   & 22.85 \\ \hline
			\end{tabular}
		}
		\label{tbl:results_hbb}
	\end{table*}

	\subsection{Top 3 Submissions on the Task1}
	\textbf{1st Place.} The \textbf{Alibaba\_AI\_Earth}, team of \textit{Qiang Zhou, Chaohui Yu} from \textit{Alibaba Group}, adopt the ReDet~\cite{ReDet} framework as baseline for this challenge. Then they optimize the baseline in mainly two ways, \ie, single model optimization and test-time augmentation (TTA). Specifically, as for single model optimization, they apply Swin~\cite{swin} transformer as the backbone, Guided Anchor~\cite{wang2019region} as the RPN, Double-Head~\cite{wu2020rethinking} as the detection head. In addition, they incorporate RiRoI Align to GRoIE~\cite{rossi2020novel} to achieve a better single model. They further use SWA~\cite{swa} to train each single model for an extra 12 epochs using cyclical learning rates and then average these 12 checkpoints as the final model. Besides, they use random rotation, random crop, random flip (vertical and horizontal), and multi-scale training as data augmentation in the training phase. As for TTA, they apply multi-scale testing by resizing the images by factors of $[0.5, 1.0, 1.5]$ and then crop the images into patches of size 1,024$\times$1,024. Since the performance of different models in each category may be different, to leverage the advantages of different single models, they further perform model ensembling to achieve better performance. The OBB results and HBB results are shown in Table~\ref{tbl:results_rbb} and Table~\ref{tbl:results_hbb}, respectively.

	\textbf{2nd Place.} The \textbf{Tsinghua\_cc}, team of \textit{Chenchen Fan} from \textit{Tsinghua}, used the official ReDet~\cite{ReDet}~\footnote{\url{https://github.com/csuhan/ReDet}} as their baseline and then improve performance based on the baseline. As for the data process, since they use multi-scale training and multi-scale testing, they prepare the train/val dataset, and the test challenge dataset with scales $[0.5, 0.75, 1.0, 1.5]$, and then resize the cropped patch to 1024x1024. As for training, they train multiple single models (backbone includes ReR50, res50, res101) with 12 epochs or 24 epochs. Then they use SWA~\cite{swa} training to improve the performance of each single model. As for testing, they use multi-scale testing and rotate testing. Finally, they ensemble the results of multiple single models.
	
	\textbf{3rd Place.} The \textbf{HIK\_HOW}, team of \textit{Kaixuan Hu, Yingjia Bu, Wenming Tan} from \textit{Hikvision Research Institute}, apply Oriented Reppoints~\cite{oreppoints} and ROI Transformer~\cite{RoITransformer} as baseline-detectors.
	They only use the competition dataset for training using pre-trained models from ImageNet, and no extra data was added in training. About backbone, they found that Swin Transformer~\cite{swin} + Path Aggregation Network (PANet)~\cite{PANet} could achieve better performance than ResNet~\cite{resnet}. Another issue with ODAI is that the hyperparameter setting of conventional object detectors learned from natural images is not appropriate for aerial images due to domain differences. Thus they changed some hyperparameters such as anchor scales and anchor ratios. Besides, tricks like model ensembling, data augmentation, test time augmentation are also adopted to improve the final performance.

	\subsection{Top 3 Submissions on the Task2}
	\textbf{Alibaba\_AI\_Earth}, \textbf{HIK\_HOW}, and \textbf{Tsinghua\_cc} ranked in the first, second, and third place in Task2, respectively. All three teams use the same methods described in Task1 and transfer their OBB results to HBB results for Task2. The ranking order is slightly different from that in Task1.

	\subsection{Top 3 Submissions on the Task3}
	\textbf{1st Place.} The \textbf{Alibaba\_AI\_Earth}, team of \textit{Zhe Yang,Wei Li, Shang Liu} from \textit{Alibaba Group} adopt segformer~\cite{SegFormer} and volo~\cite{VOLO} as the baseline. They cut the train and validation set to small patches for training. The patch size is 1024 $\x$ 1024, while the input image size is 896 $\x$ 896 for segformer and 512 $\x$ 512 for volo. An ensemble loss of IoU loss and CrossEntropy loss is used. An effective data augmentation pipeline including color jitter, contrast, brightness, and gaussian noise is used. In the training stage, they first use the poly learning rate policy to train the basic models and then use the cyclic learning rate policy to prepare snapshots for SWA~\cite{swa} average by using basic models as a pre-trained model. Segformer model and volo model are merged to get better performance. They find it is hard for a model to classify the large lake or river area, but the result nearby the land is reliable. Some categories are confusing between each other, for example, arbor woodland and shrub land, natural grassland, and artificial grassland. They design two specific models to distinguish the confusing category pairs.
	
	\textbf{2nd Place.} The \textbf{ZAIZ}, team of \textit{Jiaxuan Zhao, Tianzhi Ma, Zihan Gao, Lingqi Wang, and Yi Zuo, Licheng Jiao} from \textit{Xidian University}, adopt six pipelines and several weak classifiers for this challenge, and the final result is a
	fusion version of these pipelines. The three of them are based on Deeplabv3 with backbones: ResNet~\cite{resnet}, DRN~\cite{DRN}, and HRNet. Others are PSPNet~\cite{pspnet}, Decovnet, and Reco-main. 
	In the part of data preprocessing, according to the characteristics of the dataset, they choose some methods like data augmentation, multi-scale clipping, picture inversion, and Gaussian blur to enhance the data, which is difficult to discriminate. They also analyzed the variance of the dataset and standardized it before training. During training, three pipelines based on deeplabv3 are optimized by SGD. The learning rate is 2 from 0.01, the attenuation is 0.9, and the pre-training weight of cityscapes is loaded. In addition, they use TTA and Expansion Prediction during testing. Finally, they use the weighted voting to merge models and the smoothing operator to optimize the results.

	\textbf{3rd Place.} The \textbf{Go for It}, team of \textit{Chang Meng, Hao Wang, Jiahao Wang, Yiming Hui, Zhuojun Dong, Jie Zhang, Qianyue Bao, Zixiao Zhang and Fang Liu} from \textit{Key Laboratory of Intelligent Perception and Image Understanding, Ministry of Education, Xidian University}, tried different models and after conducting a lot of experiments, they finally chose a model from the deeplab series, and mainly used the Deeplabv3+~\cite{Deeplabv3+} model for data training. For data pre-processing, they used random cropping and flipping, Gaussian filtering, etc. By analyzing the submitted results, they found that the IOUs of categories like artificial grassland, shrublands, and ponds were particularly low, and the analysis of the data revealed the problem of extreme data imbalance. Therefore, they trained the above categories with low scores separately using a two-category split and improved the overall average cross-merge ratio by covering them step by step. The data cropping sizes were tried at 256, 512 and 1024, and superimposed cropping was added in the follow-up process, and the network outputs at different scales were finally superimposed and averaged for prediction. backbone was tried with ResNet-101~\cite{ResNet-101}, DRN~\cite{DRN} and Xception~\cite{Xception}. Finally they fused different data, different Backbone, different training stages of the model for hard voting. For the post-processing approach, they added CRF and TTA in a way that slightly improves on the original one.
	\subsection{Summary of Methods and Discussions}
	In the Task1, the RoI Transformer and ReDet are widely used by all the top submissions, which show the rotation-invariance is crucial in ODAI. Recent transformer based backbone (such as Swin Transformer~\cite{swin}) is also widely used and shows its advantage in ODAI. The architecture design and data augmentation involving scale and rotation are helpful in ODAI. In the Task2, most of the submissions transfer their OBB results to the HBB results. The possible reason is that the OBB results can be used to perform Rotated NMS (R-NMS)~\cite{RoITransformer,scrdet}, which is better than regular NMS for densely packed object detection. 
	{For Task3, the 1st team adopt the transformer-based segmentation network and significantly surpass the other teams. Our baseline algorithm is also a transformer-based method and outperforms most of the participants that use the CNN-based methods. The possible reason is that the transformer can significantly increase the effective receptive field~\cite{receptivefield} and obtain global information, which is very important for semantic segmentation in aerial images.}
	
	For all three tasks, Transformers are widely used, showing that it is also a promising direction in the LUAI.
	SWA~\cite{swa} is another widely used method in all three tasks to improve the generalization. However, there still exists much room to conduct research related to generalization in LUAI.
	
	\section{Conclusion}
	We organized a challenge on LUAI with three tasks of oriented object detection, horizontal object detection, and semantic segmentation. 
	In summary, this challenge received a total of 146 registrations from a worldwide range of institutes. 
	Through the results and summary, we find several promising directions in the LUAI. We hope this challenge can draw more attention from vision communities and promote future research on LUAI. 
	{\small
		\bibliographystyle{ieee_fullname}
		\bibliography{egbib}
	}
	
\end{document}